\documentclass[11pt]{article}
\usepackage{amsmath,amssymb,fullpage,graphicx}
\RequirePackage[colorlinks,citecolor=blue,urlcolor=blue, linkcolor=magenta]{hyperref}
\RequirePackage[round,sort&compress]{natbib}

\usepackage{booktabs}

\usepackage{setspace}

\usepackage{tikz}

\usetikzlibrary{arrows.meta, positioning}

\usepackage{mathptm}
\usepackage{subfigure}
\let\hat\widehat
\let\tilde\widetilde

\newcommand{\E}{\mbox{$\mathbb{E}$}}
\newcommand{\R}{\mbox{$\mathbb{R}$}}

\title{A Frequentist Statistical Introduction to Variational Inference, Autoencoders, and Diffusion Models}
\author{Yen-Chi Chen}
\date{\today} 

\begin{document}

\maketitle

\begin{abstract}
While Variational Inference (VI) is central to modern generative models like Variational Autoencoders (VAEs) and Denoising Diffusion Models (DDMs), its pedagogical treatment is split across disciplines. In statistics, VI is typically framed as a Bayesian method for posterior approximation. In machine learning, however, VAEs and DDMs are developed from a Frequentist viewpoint, where VI is used to approximate a maximum likelihood estimator. This creates a barrier for statisticians, as the principles behind VAEs and DDMs are hard to contextualize without a corresponding Frequentist introduction to VI.
This paper provides that introduction: we explain the theory for VI, VAEs, and DDMs from a purely Frequentist perspective, starting with the classical Expectation-Maximization (EM) algorithm. We show how VI arises as a scalable solution for intractable E-steps and how VAEs and DDMs are natural, deep-learning-based extensions of this framework, thereby bridging the gap between classical statistical inference and modern generative AI.

\end{abstract}

{
\begin{spacing}{0}
\tableofcontents
\end{spacing}
}

\section{Introduction}

Variational Inference (VI) is a powerful set of methods in modern machine learning. In the statistical literature, however, VI is most commonly introduced within a Bayesian framework, where it serves as an indispensable tool for approximating intractable posterior distributions \citep{bishop2006pattern, blei2017variational,sjolund2023tutorial, kejzlar2024introducing}.

Paradoxically, two of VI's most successful applications, the Variational Autoencoder (VAE) and the Denoising Diffusion Model (DDM), are typically constructed from a Frequentist perspective. Influential tutorials on VAEs \citep{doersch2016vae,kingma2019introduction} and DDMs \citep{luo2022understanding, chan2024tutorial} do not place priors on the model parameters. Instead, their goal is to approximate the maximum likelihood estimator (MLE) for a complex generative model\footnote{Early works on applying VI to graphical models are also based on this frequentist perspective although the use of VI in graphical models is slightly different from the current VI; see \cite{jordan1999variational,wainwright2008graphical}. }. This methodological divergence has created a pedagogical gap: while VAEs and DDMs are central to AI, their adoption in the statistics community has been slower, partly due to the lack of an introduction that frames these methods in a way that is natural for many statisticians.

This paper aims to fill this critical gap. We provide a self-contained introduction to VI, VAEs, and DDMs grounded entirely in Frequentist principles. By demonstrating that these techniques are fundamentally powerful algorithms for optimization and function approximation \citep{ormerod2010explaining,chen2018bootstrapVI}, independent of a Bayesian context, we hope to make these powerful generative models more accessible and intuitive for the statistics community.

\emph{Outline.} 
We begin in Section \ref{sec::latent} by establishing a foundation in Frequentist latent variable models and reviewing the Expectation-Maximization (EM) algorithm. We focus on two key variants--the Monte Carlo EM (MCEM) algorithm and the regularized Q-function--that directly motivate the transition to Variational Inference (VI).
Building on this, Section \ref{sec::VI} introduces VI as a general method for approximating the intractable E-step of the EM algorithm, framing the evidence lower bound (ELBO) as a variational analog to the regularized Q-function. Next, in Section \ref{sec::AVI}, we address the computational limitations of classical VI by introducing amortized VI and the Variational Autoencoder (VAE), which enable the application of VI to large-scale, deep learning models.
Finally, Section \ref{sec::DDM} presents the Denoising Diffusion Model (DDM) as a deep, hierarchical extension of this same framework, composed of a forward (variational) and reverse (generative) process. We conclude our technical discussion by deriving the simplified noise-prediction objective, which is the key to the DDM's practical success as a state-of-the-art image generator.

%
%
%
%

%
%
%
%
%

\section{Latent Variable Model}	\label{sec::latent}

Suppose our data are i.i.d. random variables $X_1,\cdots, X_n \sim p_0$, where $p_0$
is some unknown PDF and each $X_i\in\R^d$.
A standard parametric approach assumes a model on $p_0$ and the statistical task is to
estimate the underlying parameter.

However, conventional models such as Gaussian are often too simple to adequately approximate the distribution well.
Therefore, we often employ latent variable models (such as mixture models) to address this issue.
Let $Z_1,\cdots, Z_n\in\R^k$ be the latent variables associated with $X_1,\cdots, X_n$. 
We then place models  $p_\theta(x|z)$ and  $p_\theta(z)$.
The quantity $\theta$ is the parameter of interest that we wish to infer from the data.
Sometimes, $p_\theta(z) = p(z)$ is a known distribution for many latent variable models 
such as factor analysis \citep{anderson2003introduction,harman1976modern}, latent trait models \citep{rasch1960probabilistic,cai2016irt,chen2021irt}, and latent space models
\citep{hoff2002latent, sewell2015dynamic}.
For simplicity, we will assume that $p(z) $ is a known distribution and does not depend on $\theta$.

In the latent variable models, the \emph{complete log-likelihood} 
$$
\ell(\theta|x,z) = \log p_\theta(x,z) = \log p_\theta(x|z) p(z)
$$
is often easy to evaluate for any given $\theta$ and $(x,z)$.
The maximization of 
$$
\ell_{n,c}(\theta) = \sum_{i=1}^n \ell(\theta|X_i,Z_i)
$$
is generally a computationally straightforward (tractable) problem if we observe both $X$ and $Z$.
Thus, estimating $\theta$ when we observe both $X,Z$ is a simple problem.

However, we do not observe $Z$, so we can only compute the \emph{observed log-likelihood}
$$
\ell(\theta|x) = \log p_\theta(x) = \log \int p_{\theta}(x,z)dz
$$
rather than the complete log-likelihood.
Under the observed log-likelihood,
the maximum likelihood estimator (MLE) is
$$
\hat \theta_{MLE} = {\sf argmax}_\theta \ell_n(\theta) = {\sf argmax}_\theta \sum_{i=1}^n \ell(\theta|X_i).
$$
Unfortunately, due to the integral in $\ell(\theta|x)$,
$$
\ell(\theta |x) = \log p_\theta(x) = \log \int p_\theta(x,z)dz,
$$
computing $\hat \theta_{MLE}$ is generally computationally challenging (intractable).
To resolve this issue, statisticians often use the  EM algorithm.

\subsection{EM algorithm}

The expectation-maximization (EM) algorithm \citep{dempster1977em}
starts with an initial point $\theta^{(0)}$ and creates a sequence $\theta^{(1)},\theta^{(2)},\cdots,$
via the following two steps (E-step and M-step) for each $t=0,1,2,3,\cdots$:
\begin{itemize}
\item {\bf E-step.}
We compute the Q-function:
\begin{equation}
Q(\theta;\theta^{(t)}|x) = \int p_{\theta^{(t)}}(z|x) \ell(\theta|x,z)dz = \E_{Z\sim p_{\theta^{(t)}}(z|x)}[\ell(\theta|x,Z)|X=x].
\label{eq::Q}
\end{equation}

\item {\bf M-step.}
We update the parameter $\theta$ via
$$
\theta^{(t+1)} = {\sf argmax}_\theta\quad Q_n(\theta;\theta^{(t)}), \qquad Q_n(\theta;\theta^{(t)}) =  \sum_{i=1}^n Q(\theta;\theta^{(t)}|X_i).
$$

\end{itemize}

In other words, the EM algorithm is essentially replacing the direct maximization of the intractable log-likelihood function $\ell_n(\theta)$
with the iterative maximization of a more tractable  Q-function $Q_n(\theta;\theta^{(t)})$. 

It is known that the EM algorithm has a non-decreasing property \citep{wu1983em}:
\begin{equation}
\ell_n(\theta^{(t+1)}) \geq \ell_n(\theta^{(t)}).
\label{eq::EM}
\end{equation}
Thus, running the EM algorithm 
is guaranteed to not decrease the likelihood value, though it may converge to a local, rather than global, maximum.

\subsection{MCEM: Monte Carlo EM} \label{sec::MCEM}

When the integral in the E-step (equation \eqref{eq::Q}) is intractable, a common solution is to approximate the Q-function using Monte Carlo integration. This approach is known as the \emph{Monte Carlo EM (MCEM)} algorithm \citep{wei1990mcem}.

The rationale is straightforward. We know that if both $(X,Z)$ were observed, the complete-data log-likelihood maximization would be tractable. A simple Monte Carlo approximation of the E-step, therefore, involves generating a single realization
$$
\tilde Z \sim p_{\theta^{(t)}}(z|x)
$$
and using its complete log-likelihood
$$
\tilde Q(\theta;\theta^{(t)}|x) = \ell(\theta|x,\tilde Z)
$$
as a stochastic approximation to the true Q-function. When applying this to the full dataset, we would generate a single latent variable $\tilde Z_i$ from $p_{\theta^{(t)}}(z|X_i)$ for each observation $X_i$. The M-step then reduces to a conventional MLE problem for the complete data $(X_1,\tilde Z_1),\dots, (X_n,\tilde Z_n)$, which is computationally straightforward.

To reduce the Monte Carlo error from this single-realization approximation, the standard MCEM algorithm generates multiple realizations
\begin{equation}
\tilde{Z}^{(1)},\dots, \tilde{Z}^{(M)} \sim p_{\theta^{(t)}}(z|x)
\label{eq::MCEM}
\end{equation}
and uses their average to form a more stable approximation to the Q-function:
$$
\tilde Q_M(\theta;\theta^{(t)}|x) = \frac{1}{M}\sum_{m=1}^M \ell(\theta|x,\tilde{Z}^{(m)}).
$$
By the law of large numbers, as $M\to\infty$, this Monte Carlo approximation $\tilde Q_M(\theta;\theta^{(t)}|x)$ converges to the true Q-function. Thus, MCEM provides a general method for approximating the E-step when it cannot be computed analytically. From a missing data perspective \citep{little2019missing}, MCEM can be viewed as a multiple imputation method: we are imputing the latent variables multiple times and using all the imputed results together for the M-step.

\subsection{Regularization form of the Q-function}

In the EM algorithm, the Q-function is central to the entire process. While it can be understood from a missing data perspective, an alternative and powerful view frames it as a regularized log-likelihood function \citep{neal1998view}.

Recall that the standard Q-function is
$$
Q(\theta;\theta^{(t)}|x) = \int p_{\theta^{(t)}}(z|x) \ell(\theta|x,z)dz.
$$
Maximizing this Q-function with respect to $\theta$ is equivalent to maximizing the following objective:
\begin{equation}
Q^*(\theta;\theta^{(t)}|x) = Q(\theta;\theta^{(t)}|x) - \int p_{\theta^{(t)}}(z|x) \log p_{\theta^{(t)}}(z|x) dz,
\label{eq::Q1}
\end{equation}
since the second term, the negative entropy of $p_{\theta^{(t)}}(z|x)$, does not depend on $\theta$. Thus, we can rewrite the M-step as
$$
\theta^{(t+1)} = {\sf argmax}_\theta\ Q^*_n(\theta;\theta^{(t)}),
$$
where $ Q^*_n(\theta;\theta^{(t)}) = \sum_{i=1}^n Q^*(\theta;\theta^{(t)}|X_i)$.

This modified Q-function, $Q^*$, has an insightful decomposition:
\begin{equation}
\begin{aligned}
Q^*(\theta;\theta^{(t)}|x) &= \int p_{\theta^{(t)}}(z|x) \ell(\theta|x,z)dz - \int p_{\theta^{(t)}}(z|x) \log p_{\theta^{(t)}}(z|x) dz\\
& = \int p_{\theta^{(t)}}(z|x) [\ell(\theta|x) + \log p_\theta(z|x)]dz - \int p_{\theta^{(t)}}(z|x) \log p_{\theta^{(t)}}(z|x) dz\\
& = \ell(\theta|x) - \int p_{\theta^{(t)}}(z|x) \log \frac{p_{\theta^{(t)}}(z|x)}{p_\theta(z|x)} dz\\
& = \ell(\theta|x) - \mathrm{KL}(p_{\theta^{(t)}}(\cdot|x)\|p_{\theta}(\cdot|x)).
\end{aligned}
\label{eq::Q2}
\end{equation}
That is,
$$
Q^*(\theta;\theta^{(t)}|x) = \ell(\theta|x) - \mathrm{KL}(p_{\theta^{(t)}}(\cdot|x)\|p_{\theta}(\cdot|x)),
$$
which can be interpreted as a regularized log-likelihood. This objective balances maximizing the log-likelihood term $\ell(\theta|x)$ with a penalty that keeps the new distribution $p_{\theta}(\cdot|x)$ close to the old one $p_{\theta^{(t)}}(\cdot|x)$. This reveals the EM algorithm as a form of proximal point algorithm \citep{neal1998view}.

More explicitly, the M-step is equivalent to a penalized log-likelihood maximization:
\begin{equation}
\begin{aligned}
\theta^{(t+1)} &= {\sf argmax}_\theta\ Q^*_n(\theta;\theta^{(t)})\\
& = {\sf argmax}_\theta\ \left\{ \ell_n(\theta) - \sum_{i=1}^n\mathrm{KL}(p_{\theta^{(t)}}(\cdot|X_i)\|p_{\theta}(\cdot|X_i)) \right\}\\
& = {\sf argmin}_\theta\ \left\{ -\ell_n(\theta) + \sum_{i=1}^n\mathrm{KL}(p_{\theta^{(t)}}(\cdot|X_i)\|p_{\theta}(\cdot|X_i)) \right\}.
\end{aligned}
\label{eq::M2}
\end{equation}

Equation \eqref{eq::M2} also provides a direct proof of the non-descending property of EM. Since the KL divergence is non-negative and is $0$ only when the two distributions are identical, we have:
\begin{align*}
Q^*_n(\theta^{(t)};\theta^{(t)}) &= \ell_n(\theta^{(t)}) - 0 = \ell_n(\theta^{(t)}) \\
Q^*_n(\theta^{(t+1)};\theta^{(t)}) &= \ell_n(\theta^{(t+1)}) - \sum_{i=1}^n\mathrm{KL}(p_{\theta^{(t)}}(\cdot|X_i)\|p_{\theta^{(t+1)}}(\cdot|X_i)).
\end{align*}
By the definition of the M-step, we know that $Q^*_n(\theta^{(t+1)};\theta^{(t)}) \geq Q^*_n(\theta^{(t)};\theta^{(t)})$. This implies:
\begin{align*}
\ell_n(\theta^{(t)}) &= Q^*_n(\theta^{(t)};\theta^{(t)})\\
& \leq Q^*_n(\theta^{(t+1)};\theta^{(t)}) \\
& = \ell_n(\theta^{(t+1)}) - \sum_{i=1}^n\mathrm{KL}(p_{\theta^{(t)}}(\cdot|X_i)\|p_{\theta^{(t+1)}}(\cdot|X_i))\\
&\leq \ell_n(\theta^{(t+1)}),
\end{align*}
which recovers the non-decreasing property from equation \eqref{eq::EM}.

\subsection{Example: limitation of the EM Algorithm} \label{sec::limit}

While the EM algorithm is an effective method when the MLE has no closed-form solution, its applicability is limited by the tractability of the E-step. Here, we present an example to illustrate this limitation.

Let $X_1,\dots,X_n\in\mathbb{R}^{d}$ be i.i.d. continuous random variables representing our data, and let $Z_i\in \mathbb{R}^k$ be the corresponding latent variables such that both $d,k$ are high-dimensional. We model the PDF of $X$ using a latent variable model:
$$
X|Z \sim N(\mu_\theta(Z), \sigma^2_\theta(Z) \mathbf{I}_d),\qquad Z\sim N(0,\mathbf{I}_k),
$$
where $\mu_{\theta}: \mathbb{R}^k\to\mathbb{R}^d$ and $\sigma^2_\theta:\mathbb{R}^k\to\mathbb{R}$ are functions parameterized by $\theta$. One can think of $\mu_\theta(z)$ and $\sigma^2_\theta(z)$ as neural network models.

Clearly, the marginal log-likelihood,
$$
\ell(\theta|x) = \log \int (2\pi \sigma^2_\theta(z))^{-d/2} \exp\left(-\frac{\|x-\mu_\theta(z)\|^2}{2\sigma^2_\theta(z)}\right) \cdot (2\pi)^{-k/2} \exp\left(-\frac{1}{2}\|z\|^2\right)dz,
$$
is intractable, as it involves a high-dimensional integral over $z$. While one could use Monte Carlo integration to approximate it, a very large number of samples would be required. This is because the region of high density for $p_\theta(x|z)$ as a function of $z$ generally has little overlap with the typical set of the distribution $p(z)$, making naive importance sampling from the $p(z)$ highly inefficient. This problem is particularly severe in modern machine learning, where the dimensions of $X$ and $Z$ can be in the millions or billions for applications like image generation \citep{rombach2022high, saharia2022photorealistic}.

Alternatively, we might consider the EM algorithm. However, the E-step requires computing the  distribution:
\begin{equation}
p_\theta(z|x) = \frac{p_\theta(x,z)}{\int p_\theta(x,z')dz'} = \frac{\sigma^{-d}_\theta(z) \exp\left(-\frac{\|x-\mu_\theta(z)\|^2}{2\sigma^2_\theta(z)}\right) \cdot \exp\left(-\frac{1}{2}\|z\|^2\right)}{\int \sigma^{-d}_\theta(z') \exp\left(-\frac{\|x-\mu_\theta(z')\|^2}{2\sigma^2_\theta(z')}\right) \cdot \exp\left(-\frac{1}{2}\|z'\|^2\right)dz'}.
\label{eq::ex1}
\end{equation}
In general, this distribution does not belong to any standard distributional family, making the analytical computation of the Q-function in equation \eqref{eq::Q} intractable.

If we resort to the MCEM approach, sampling from the complex distribution in equation \eqref{eq::ex1} is also a non-trivial problem. While Markov chain Monte Carlo (MCMC) methods might work for small $d$ and $k$, they become prohibitively slow when these dimensions are large, as is common in high-dimensional settings like image generation.

\section{Variational Approximation} \label{sec::VI}

The example in Section \ref{sec::limit} highlights a central challenge in complex latent variable models: the  distribution $p_\theta(z|x)$ is often intractable, making both exact inference and sampling difficult. Variational Inference (VI; Chapter 10 of \citealt{bishop2006pattern}) provides a powerful framework for resolving this issue. The core idea of VI is to approximate the intractable  $p_\theta(z|x)$ with a tractable variational distribution, $q_\omega(z)$, chosen from a family of distributions parameterized by $\omega$ (e.g., a multivariate Gaussian). With this tractable approximation, we can then derive a new objective function analogous to the Q-function.

The VI objective is derived by constructing a lower bound on the log-likelihood function:
\begin{align*}
\ell(\theta|x) & = \log p_\theta(x) = \log \int p_\theta(x,z)dz \\
& = \log \int \frac{p_\theta(x,z)}{q_\omega(z)} q_{\omega}(z)dz \\
& \geq \int q_{\omega}(z) \log \frac{p_{\theta}(x,z)}{q_{\omega}(z)}dz \qquad \text{(Jensen's inequality)} \\
& = \int q_\omega(z) \log p_\theta(x,z) dz - \int q_{\omega}(z) \log q_\omega(z)dz \\
& = \int q_\omega(z) \ell(\theta|x,z) dz + H(q_\omega) \\
&\equiv {\sf ELBO}(\theta,\omega|x),
\end{align*}
where $H(q_\omega)$ is the entropy of the variational distribution $q_\omega$. The quantity ${\sf ELBO}(\theta,\omega|x)$ is called the \emph{evidence lower bound (ELBO)}. Note that for this bound to be valid, the support of $q_\omega(z)$ must contain the support of $p_\theta(z|x)$.

This ELBO bears a strong resemblance to the modified Q-function, $Q^*$, from equation \eqref{eq::Q1}:
\begin{equation}
\begin{aligned}
{\sf ELBO}(\theta,\omega|x) & = \int q_\omega(z) \ell(\theta|x,z) dz + H(q_\omega), \\
Q^*(\theta;\theta^{(t)}|x) &= \int p_{\theta^{(t)}}(z|x) \ell(\theta|x,z)dz + H(p_{\theta^{(t)}}(\cdot|x)).
\end{aligned}
\label{eq::EQ}
\end{equation}
Essentially, VI replaces the true (but intractable)  $p_{\theta^{(t)}}(z|x)$ in the EM objective with the tractable variational distribution $q_\omega(z)$.

Furthermore, we can rewrite the ELBO using the same decomposition as in equation \eqref{eq::Q2}:
\begin{align*}
{\sf ELBO}(\theta,\omega|x) & = \int q_\omega(z) [\ell(\theta|x) + \log p_\theta(z|x)] dz + H(q_\omega) \\
& = \ell(\theta|x) - \mathrm{KL}(q_{\omega}(\cdot)\| p_\theta(\cdot|x)).
\end{align*}
This form reveals that maximizing the ELBO with respect to $\omega$ is equivalent to minimizing the KL-divergence between the variational distribution and the true conditional distribution. This makes the goal of VI explicit: choose $\omega$ such that $q_{\omega}(z) \approx p_\theta(z|x)$. Since the target of our approximation, $p_\theta(z|x)$, depends on the observation $x$, the optimal variational distribution must also depend on $x$. This motivates assigning a unique variational parameter, $\omega_i$, to each data point, $X_i$.

Therefore, for a dataset $X_1,\dots,X_n$, the total ELBO is
$$
{\sf ELBO}(\theta,\omega_1,\dots, \omega_n) = \sum_{i=1}^n {\sf ELBO}(\theta,\omega_i|X_i),
$$
and the VI estimators are found by a joint maximization:
\begin{equation}
(\hat \theta_{VI}, \hat \omega_1,\dots,\hat\omega_n) = {\sf argmax}_{\theta,\omega_1,\dots, \omega_n}\ \sum_{i=1}^n {\sf ELBO}(\theta,\omega_i|X_i).
\label{eq::VI}
\end{equation}
This optimization can also be viewed as a nested procedure. Let
$$
\omega^*(x;\theta) = {\sf argmax}_{\omega} {\sf ELBO}(\theta,\omega|x) = {\sf argmin}_{\omega}\ \mathrm{KL}(q_{\omega}(\cdot)\| p_\theta(\cdot|x))
$$
be the optimal choice of $\omega$ for a given $\theta$ and $x$. Then the estimator for $\theta$ can be written as:
\begin{equation}
\hat \theta_{VI} = {\sf argmax}_{\theta}\ \sum_{i=1}^n {\sf ELBO}(\theta,\omega^*(X_i;\theta)|X_i).
\label{eq::VI2}
\end{equation}

In certain conjugate models, such as Latent Dirichlet Allocation \citep{blei2003latent}, the optimal $\omega^*(x;\theta)$ has a closed-form solution or can be found via an efficient iterative procedure like CAVI \citep{blei2017variational}. In such cases, equation \eqref{eq::VI2} can be optimized in a manner similar to the EM algorithm. However, for the general class of models considered in Section \ref{sec::limit}, $\omega^*(x;\theta)$ does not have a closed-form solution. We must then resort to numerical methods, such as the gradient ascent method \citep{boyd2004convex, bubeck2015convex}, which we detail in the next section.

\subsection{Gradient of the ELBO and the reparameterization trick}	\label{sec::repara}

The optimization for VI differs from a standard gradient ascent because the optimal variational parameters $\omega_i$ depend on the global parameters $\theta$. This coupling necessitates a nested or alternating optimization scheme.

Here, we summarize a gradient ascent procedure to compute the VI estimators, which can be easily modified into a stochastic gradient ascent algorithm \citep{hoffman2013stochastic}. We start with an initial value $\theta^{(0)}$ and then iterate the following steps until convergence:

For a given $\theta^{(t)}$, we first find the optimal variational parameters for each observation by running an inner loop of gradient ascent. For each $i=1,\dots,n$, we find $\tilde\omega^{(t)}_i$ by initializing at $\omega^{(0)}_i$ (often using a warm start, $\omega^{(0)}_i = \tilde\omega^{(t-1)}_i$) and iterating:
\begin{equation}
\omega^{(s+1)}_i = \omega^{(s)}_i + \gamma_{\omega}  \nabla_{\omega_i}{\sf ELBO}(\theta^{(t)},\omega^{(s)}_i|X_i),
\label{eq::GD::omega}
\end{equation}
where $\gamma_\omega>0$ is a stepsize. The convergent point, $\tilde\omega^{(t)}_i \approx \omega^*(X_i;\theta^{(t)})$, is the optimal variational parameter for observation $X_i$ under the current global model $\theta^{(t)}$.

After updating all the local variational parameters, we perform a single gradient ascent step on the global parameters:
\begin{equation}
\theta^{(t+1)} = \theta^{(t)} + \gamma_\theta  \sum_{i=1}^n \nabla_\theta{\sf ELBO}(\theta^{(t)},\tilde\omega^{(t)}_i|X_i),
\label{eq::GD::theta}
\end{equation}
where $\gamma_\theta>0$ is a stepsize. This entire process is iterated until convergence. The reason for this nested structure is that if we update $\theta^{(t)}$ to $\theta^{(t+1)}$, the previous variational parameters $\tilde\omega^{(t)}_i$ are no longer the best approximation to the new distribution $p_{\theta^{(t+1)}}(z|X_i)$, so they must be re-optimized.

\bigskip
\noindent\textbf{Gradient with respect to $\theta$.}
We now provide details on computing the gradient $\nabla_\theta{\sf ELBO}(\theta,\omega_i|X_i)$. The second term in the ELBO definition (equation \eqref{eq::EQ}), the entropy, does not depend on $\theta$. Thus, the gradient is:
\begin{align*}
\nabla_\theta{\sf ELBO}(\theta,\omega_i|X_i)& = \nabla_\theta\int q_{\omega_i}(z) \ell(\theta|X_i,z) dz \\
& = \int q_{\omega_i}(z) \nabla_\theta \ell(\theta|X_i,z) dz \\
& = \int q_{\omega_i}(z) s(\theta|X_i,z) dz \\
& = \mathbb{E}_{Z\sim q_{\omega_i}}[s(\theta|X_i, Z)],
\end{align*}
where $s(\theta|x,z) = \nabla_\theta \ell(\theta|x,z)= \nabla_\theta \log p_\theta(x,z)$  is the complete-data score function. Since we can easily sample from the variational distribution $q_{\omega_i}$, this expectation can be approximated via Monte Carlo integration. We generate $\tilde Z^{(1)},\dots, \tilde Z^{(M)} \sim q_{\omega_i}$ and then compute the gradient estimate:
\begin{equation}
\tilde{\nabla_\theta{\sf ELBO}}(\theta,\omega_i|X_i) = \frac{1}{M} \sum_{m=1}^M s(\theta|X_i, \tilde Z^{(m)}).
\label{eq::MC::theta}
\end{equation}
This approach is analogous to how MCEM approximates the gradient of the Q-function. In VI, this Monte Carlo average is used to numerically approximate the gradient of the ELBO. The crucial advantage over MCEM is that we sample from the tractable variational distribution $q_{\omega_i}$ instead of the intractable $p_\theta(z|X_i)$, thus avoiding the primary computational bottleneck.

\bigskip
\noindent\textbf{Gradient with respect to $\omega_i$ and the reparameterization trick.}
We now consider the gradient with respect to the variational parameters, $\omega_i$, which is essential for the update step in equation \eqref{eq::GD::omega}. Both terms in the ELBO depend on $\omega_i$:
\begin{equation}
\nabla_{\omega_i}{\sf ELBO}(\theta,\omega_i|X_i) = \nabla_{\omega_i}\int q_{\omega_i}(z) \ell(\theta|X_i,z) dz + \nabla_{\omega_i} H(q_{\omega_i}),
\label{eq::GD::omega1}
\end{equation}
where $H(q_{\omega_i}) = -\int q_{\omega_i}(z) \log q_{\omega_i}(z)dz$ is the entropy of the variational distribution. For many standard distributions, the gradient of the entropy term, $\nabla_{\omega_i} H(q_{\omega_i})$, can be computed analytically. The main challenge, therefore, lies in computing the gradient of the first term.

To make this gradient tractable, we must choose a convenient variational family. A common and powerful choice is the \emph{Gaussian mean-field} family. Specifically, we assume $q_{\omega_i}(z)$ follows a multivariate Gaussian distribution with a diagonal covariance matrix, $N(\alpha_i, {\sf diag}(\beta_i^2))$, where the variational parameters are $\omega_i = (\alpha_i, \beta_i) \in \mathbb{R}^k \times \mathbb{R}^k_{>0}$. Here, $\alpha_i$ is the mean vector and $\beta_i$ is the vector of standard deviations.
The Gaussian mean-field distribution is a multivariate Gaussian with independent coordinates.


This choice enables the use of the \emph{reparameterization trick}. A random variable $Z \sim N(\alpha_i, {\sf diag}(\beta_i^2))$ can be expressed as a deterministic transformation of its parameters and a standard normal random variable $\epsilon \sim N(0, \mathbf{I}_k)$:
$$
Z = \alpha_i + \beta_i \odot \epsilon,
$$
where $\odot$ denotes the element-wise product. This allows us to rewrite the expectation so that the gradient can be passed inside the integral:
$$
\nabla_{\omega_i}\int q_{\omega_i}(z) \ell(\theta|X_i,z) dz = \int p_E(\epsilon) \nabla_{\omega_i}\ell(\theta|X_i, \alpha_i +\beta_i \odot\epsilon)d\epsilon,
$$
where $p_E$ is the PDF of $N(0, \mathbf{I}_k)$. The gradient with respect to $\alpha_i$ is then
\begin{align*}
\nabla_{\alpha_i}\mathbb{E}_{Z\sim q_{\omega_i}}[\ell(\theta|X_i,Z)] &= \int p_E(\epsilon) \nabla_{\alpha_i} \ell(\theta|X_i, \alpha_i +\beta_i \odot\epsilon)d\epsilon \\
&= \mathbb{E}_{\epsilon\sim p_E}[\nabla_{z} \ell(\theta|X_i, z)|_{z = \alpha_i +\beta_i \odot\epsilon}].
\end{align*}
This expectation can be approximated with a Monte Carlo estimate. By generating $\epsilon^{(1)},\dots, \epsilon^{(M)} \sim N(0, \mathbf{I}_k)$, we have:
$$
\tilde{\nabla_{\alpha_i}\mathbb{E}_{Z\sim q_{\omega_i}}}[\ell(\theta|X_i,Z)] = \frac{1}{M} \sum_{m=1}^M \nabla_{z} \ell(\theta|X_i, \alpha_i +\beta_i \odot\epsilon^{(m)}).
$$
A similar derivation for $\beta_i$ yields the Monte Carlo estimate:
$$
\tilde{\nabla_{\beta_i}\mathbb{E}_{Z\sim q_{\omega_i}}}[\ell(\theta|X_i,Z)] = \frac{1}{M} \sum_{m=1}^M \epsilon^{(m)} \odot \nabla_{z} \ell(\theta|X_i, \alpha_i +\beta_i \odot\epsilon^{(m)}).
$$
Combining these with the analytical gradient of the entropy term (for an isotropic Gaussian, $\nabla_{\alpha_{ij}}H(q_{\omega_i}) = 0$ and $\nabla_{\beta_{ij}}H(q_{\omega_i}) = 1/\beta_{ij}$), we can efficiently compute the full gradient $\nabla_{\omega_i}{\sf ELBO}$ and perform the gradient ascent step in equation \eqref{eq::GD::omega}.

\subsubsection{Conditions for fast gradient ascent}\label{sec::conditions}
The above derivation highlights two key conditions for efficient, gradient-based variational inference:
\begin{itemize}
    \item \textbf{Differentiable Model.} The complete-data log-likelihood $\ell(\theta|x,z) = \log p_\theta(x,z)$ must be differentiable with respect to both the model parameters $\theta$ and the latent variables $z$. For modern deep generative models where, for instance, $X|Z=z \sim N(\mu_\theta(z), \Sigma_\theta(z))$, this requires that the functions $\mu_\theta(z)$ and $\Sigma_\theta(z)$ are differentiable. This condition is readily met by neural networks, where these gradients are computed efficiently via the backpropagation algorithm used in modern automatic differentiation frameworks \citep{rumelhart1986learning, baydin2018automatic}.

    \item \textbf{Reparameterizable Variational Family.} The variational distribution $q_\omega(z)$ must be reparameterizable. Many common continuous distributions satisfy this property, often via the inverse CDF method where a sample can be generated as $Z = F_\omega^{-1}(U)$ for $U\sim {\sf Uniform}[0,1]$. This allows the gradient $\nabla_\omega$ to be handled effectively.
\end{itemize}

\section{Amortized Variational Inference and the Variational Autoencoder} \label{sec::AVI}

The VI framework described previously has two main limitations. First, it requires optimizing $n$ distinct variational parameters, $(\omega_1, \dots, \omega_n)$, which becomes computationally expensive as the sample size $n$ grows. Second, it is conceptually awkward to approximate a conditional distribution $p_\theta(z|X_i)$ using a marginal distribution $q_{\omega_i}(z)$.

\emph{Amortized Variational Inference (AVI; \citealt{gershman2014amortized})} resolves both issues by replacing the separate variational distributions with a single, conditional inference model, $q_\phi(z|x)$. Here, the variational parameters $\phi$ are shared across all data points. This way, we only need to optimize one set of parameters, regardless of sample size. The celebrated Variational Autoencoder (VAE; \citealt{kingma2014vae}) is a prominent application of AVI, particularly for image data. 

The variational distribution in AVI, $q_\phi(z|x)$, may be constructed from a non-amortized variational distribution $q_\omega(z)$ via modeling $\omega = f_\phi(x)$ for some function $f$, generally a neural network model. In this construction, $q_\phi(z|x) = q_{\omega = f_\phi(x)}(z)$. Section \ref{sec::gaussian} provides an example of this.


Under AVI, the ELBO is derived similarly:
\begin{align*}
\ell(\theta|x) & = \log p_\theta(x) = \log \int \frac{p_\theta(x,z)}{q_\phi(z|x)} q_{\phi}(z|x)dz\\
& \geq \int q_{\phi}(z|x) \log \frac{p_{\theta}(x,z)}{q_{\phi}(z|x)}dz \qquad \text{(Jensen's inequality)}\\
& = \int q_{\phi}(z|x) \ell(\theta|x,z) dz + H(q_\phi(\cdot|x)) \\
&\equiv {\sf ELBO}_A(\theta,\phi|x).
\end{align*}
Comparing the objectives highlights the progression from EM to AVI:
\begin{equation}
\begin{aligned}
{\sf ELBO}_A(\theta,\phi|x) & = \mathbb{E}_{Z\sim q_\phi(z|x)}[\ell(\theta|x,Z)] + H(q_\phi(\cdot|x)),\\
{\sf ELBO}(\theta,\omega|x) & = \mathbb{E}_{Z\sim q_\omega(z)}[\ell(\theta|x,Z)] + H(q_\omega(\cdot)),\\
Q^*(\theta;\theta^{(t)}|x) &= \mathbb{E}_{Z\sim p_{\theta^{(t)}}(z|x)}[\ell(\theta|x,Z)] + H(p_{\theta^{(t)}}(\cdot|x)).
\end{aligned}
\label{eq::EQ2}
\end{equation}
This makes it clear that AVI approximates the true distribution $p_{\theta^{(t)}}(z|x)$ with a conditional variational distribution $q_\phi(z|x)$. The regularization form of the ELBO,
$$
{\sf ELBO}_A(\theta,\phi|x) = \ell(\theta|x) - \mathrm{KL}(q_{\phi}(\cdot|x)\| p_\theta(\cdot|x)),
$$
confirms that the optimal $\phi$ is the one that minimizes the KL divergence. Since both $q_\phi$ and $p_\theta$ are conditional on $x$, a single shared parameter vector $\phi$ is sufficient for all $n$ samples.

Thus, the AVI estimator is found by a joint maximization over a fixed number of parameters:
\begin{equation}
(\hat \theta_{AVI}, \hat \phi) = {\sf argmax}_{\theta,\phi}\ \sum_{i=1}^n {\sf ELBO}_A(\theta,\phi|X_i).
\label{eq::AVI}
\end{equation}
This greatly reduces the computational complexity compared to non-amortized VI when $n$ is large. The search for the maximizer in equation \eqref{eq::AVI} is typically performed using stochastic gradient ascent.

\subsection{Example: connecting amortized and non-amortized VI}	\label{sec::gaussian}
Now we consider the specific case where our amortized variational distribution $q_\phi(z|x)$ is a Gaussian with a diagonal covariance matrix: $N(\eta_\phi(x), {\sf diag}(\delta^2_{\phi,1}(x),\dots, \delta^2_{\phi,k}(x)))$, where $\eta_\phi(x),\delta^2_\phi(x)\in\R^k$ are some functions. This is a common choice in practice and can be viewed as an amortized version of the Gaussian mean-field family from Section \ref{sec::repara}.

Recall that in the non-amortized Gaussian mean-field approach, the variational distribution for each observation $X_i$ is $q_{\omega_i}(z) = N(\alpha_i, {\sf diag}(\beta_i^2))$, where $\omega_i = (\alpha_i, \beta_i)$ is an individual parameter vector that is directly optimized.

In the amortized setting, the functions $\eta_\phi(x)$ and $\delta_\phi(x)$ (e.g., neural networks parameterized by $\phi$) are trained to predict the optimal mean and standard deviation for any given input $x$. Thus, the connection can be seen as:
$$
(\alpha_i, \beta_i) \approx (\eta_\phi(X_i), \delta_\phi(X_i)).
$$
This highlights the fundamental difference: non-amortized VI directly optimizes $n$ separate parameter vectors $(\omega_1, \dots, \omega_n)$, whereas AVI optimizes a single, global parameter vector $\phi$ for a function that generates the local parameters for each observation. While AVI greatly reduces the computational burden and allows for inference on new data points, this efficiency may come at the cost of approximation accuracy. The potential decrease in the ELBO due to the limited expressivity of the amortized function is the \emph{amortization gap} \citep{cremer2018inference, margossian2023amortized}.

\subsection{Gradient of the amortized ELBO} \label{sec::GD::AVI}

To compute the AVI estimator in equation \eqref{eq::AVI}, we can again use a gradient ascent or stochastic gradient ascent algorithm \citep{robbins1951stochastic, bottou2010large}. In AVI, the optimization is considerably simpler than in the non-amortized case because the variational parameters $\phi$ are shared across all observations. This removes the need for a nested optimization loop.

The gradient ascent is a standard procedure. Starting with initial values $\theta^{(0)}$ and $\phi^{(0)}$, the parameters are updated for $t=0,1,\dots$ until convergence:
\begin{equation}
\begin{aligned}
\theta^{(t+1)} &= \theta^{(t)}  + \gamma_\theta \nabla_\theta \sum_{i=1}^n{\sf ELBO}_{A}(\theta^{(t)},\phi^{(t)}|X_i),\\
\phi^{(t+1)} &= \phi^{(t)}  + \gamma_\phi \nabla_\phi \sum_{i=1}^n{\sf ELBO}_{A}(\theta^{(t)},\phi^{(t)}|X_i),
\end{aligned}
\label{eq::GD}
\end{equation}
where $\gamma_\theta,\gamma_\phi>0$ are stepsize parameters.

The computation of these gradients is analogous to the non-amortized case. The gradient with respect to the model parameters $\theta$ can be estimated via a Monte Carlo average, and the gradient with respect to the variational parameters $\phi$ can be efficiently computed using the reparameterization trick, assuming a suitable variational family is chosen. We provide the detailed derivations in Appendix \ref{sec::G::AVI}.

In modern applications like the VAE, it is common to specify the generative model $p_\theta(x|z)$ using a deep neural network. For instance, one might model
$$
X|Z=z \sim N(\mu_\theta(z), \Sigma_\theta(z)),
$$
where the mean and covariance functions, $\mu_\theta(z)$ and $\Sigma_\theta(z)$, are themselves parameterized by neural networks. In this setting, the required gradients of these functions with respect to both $\theta$ and $z$ can be computed efficiently via the backpropagation algorithm used in modern automatic differentiation frameworks \citep{rumelhart1986learning, baydin2018automatic}.

Thus, as long as the model is differentiable and the variational family is reparameterizable (the conditions in Section \ref{sec::conditions}), the AVI estimators can be computed efficiently via gradient ascent or stochastic gradient ascent.

\subsection{Variational Autoencoder (VAE)} \label{sec::VAE}

In a latent variable model, the data-generating process is modeled by first drawing a latent variable $Z \sim p(z)$ and then an observation $X \sim p_\theta(x|z)$. In the VAE literature, the model for the conditional distribution, $p_\theta(x|z)$, is called the \textbf{decoder}; it decodes a latent representation $Z$ into an observation $X$.

When we apply AVI, we introduce a conditional distribution $q_\phi(z|x)$ as a tractable approximation to the true conditional. This distribution can be interpreted as a model for \emph{inferring} the latent variable $Z$ from the observed variable $X$. In the VAE literature, this variational distribution $q_\phi(z|x)$ is called the \textbf{encoder}; it encodes an observation $X$ into a latent representation $Z$.

From a statistical perspective, one typically begins by specifying a scientifically-motivated generative model (the decoder, $p_\theta(x|z)$). When maximum likelihood inference for $\theta$ is difficult and the EM algorithm is intractable due to the difficulty of computing $p_\theta(z|x)$ in the E-step, we then introduce the variational distribution (the encoder, $q_\phi(z|x)$) as a computational tool for approximate inference.

The conceptual starting point, however, often differs in the deep learning literature. VAE practitioners frequently begin by designing the architecture of the encoder and then construct a corresponding decoder to model the reverse, generative mapping. The denoising diffusion models discussed in the next section exemplify this approach, where tutorials often start with the forward process (which defines the variational distribution) before deriving the reverse process (the generative model). This difference in modeling philosophy often stems from a focus on generative utility versus scientific interpretability; see Section \ref{sec::LVM} for more discussion.

To summarize the roles:
\begin{itemize}
    \item \textbf{Decoder:} The decoder, $p_\theta(x|z)$, is the model on the data-generating process.
    \item \textbf{Encoder:} The encoder, $q_\phi(z|x)$, is the variational distribution, which serves as a tractable, computational approximation to the true but intractable $p_\theta(z|x)$.
\end{itemize}

It is crucial to recognize that the decoder $p_\theta(x|z)$ and the prior $p(z)$ are sufficient to fully define the joint distribution $p_\theta(x,z)$ and, by Bayes rule, the true conditional $p_\theta(z|x)$. However, performing exact inference within this model is often intractable in high dimensions. Therefore, for computational feasibility, we introduce a separate, tractable inference model--the encoder $q_\phi(z|x)$--to approximate the true $p_\theta(z|x)$.

This implies that the encoder and decoder are, in general, incompatible. The encoder $q_\phi(z|x)$ is not the true conditional derived from the decoder and prior. Indeed, if they were compatible (i.e., if $q_\phi(z|x) = p_\theta(z|x)$), variational inference would be exact, and the EM/MCEM algorithm would be applicable. Despite this incompatibility, the encoder-decoder pairing creates a computationally feasible scheme for approximating the intractable MLE, $\hat\theta_{MLE}$, with the tractable AVI estimator, $\hat\theta_{AVI}$.

\section{Denoising Diffusion Model (DDM)} \label{sec::DDM}

\begin{figure}
\centering
\includegraphics[height=1in]{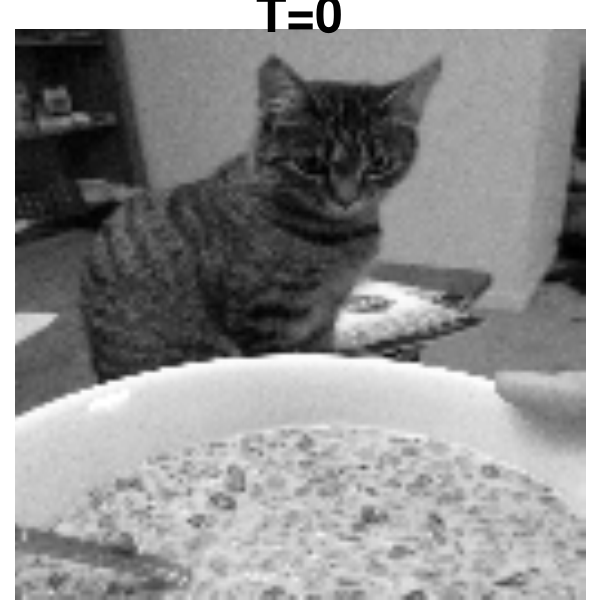}
\includegraphics[height=1in]{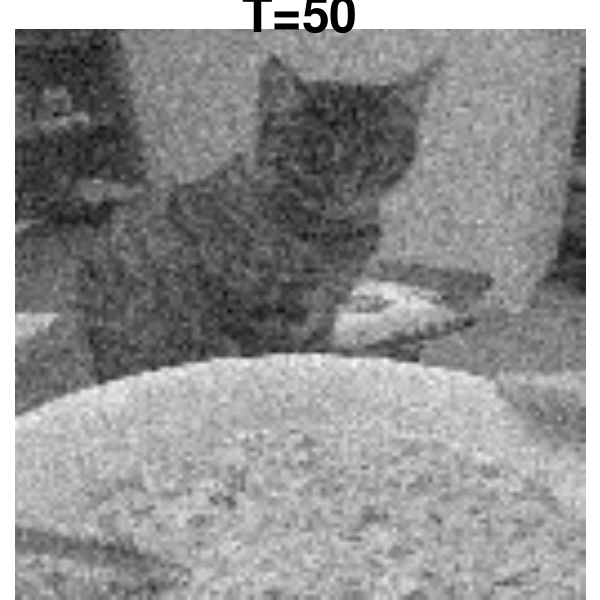}
\includegraphics[height=1in]{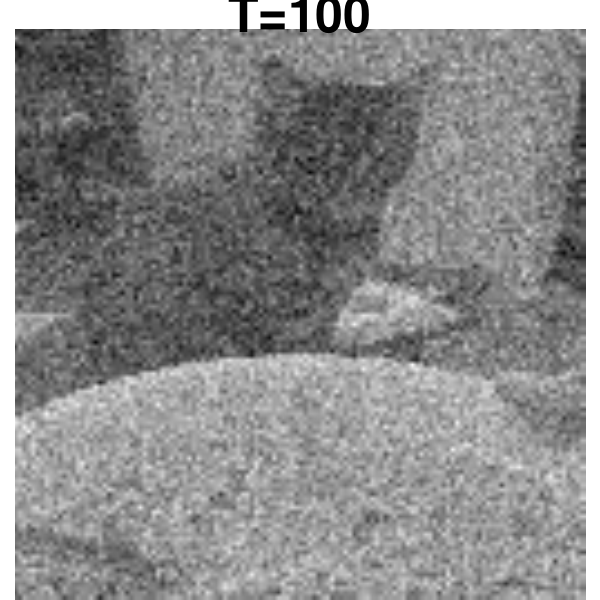}
\includegraphics[height=1in]{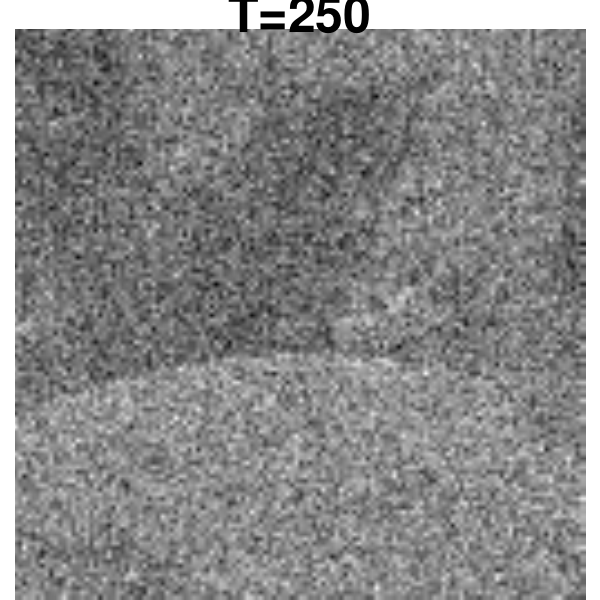}
\includegraphics[height=1in]{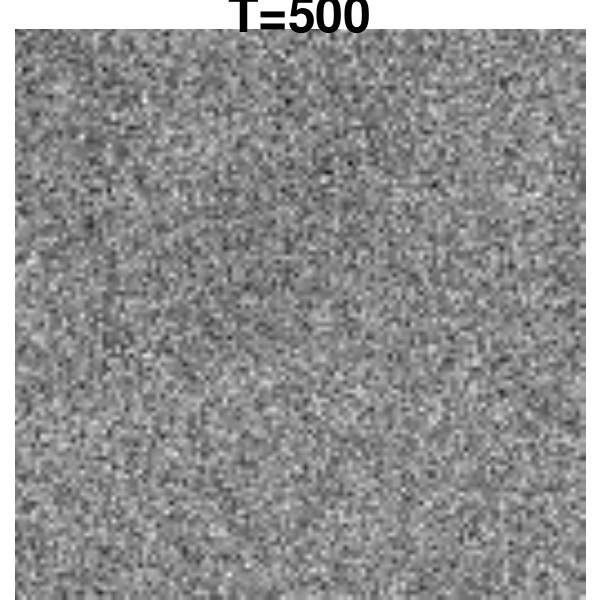}
\includegraphics[height=1in]{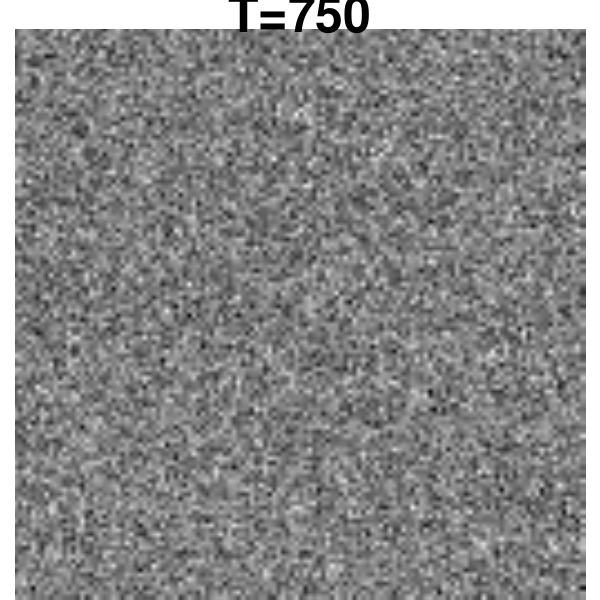}
\caption{An illustration of the DDM framework. An observation $Y_0=X_i$ is a clean image (left). The variational distribution (the \emph{forward process}) gradually adds Gaussian noise, moving from left to right. The generative model (the \emph{reverse process}) learns to reverse this, starting from noise (right) and progressively denoising it to recover a clean image.}
\label{fig::DDM::VI}
\end{figure}

The Denoising Diffusion Model (DDM), also known as a variational diffusion model, is a powerful class of generative models, particularly for image synthesis \citep{sohldickstein2015diffusion, ho2020ddpm}. The DDM can be understood as a special case of the VAE/AVI framework. Here, we frame the DDM using the language of statistical latent variable models. In short, a DDM is a deep latent variable model that is trained using an amortized variational approximation. Figure \ref{fig::DDM::VI} provides a visual summary.

\subsection{A deep latent variable model}
A conventional latent variable model is \emph{shallow}, with a single latent vector $Z$ generating an observation $X$. The DDM deepens this structure by introducing a sequence of latent variables that form a Markov chain. For simplicity, we assume all variables, both observed and latent, are of the same dimension, $X, Z \in \mathbb{R}^d$.

The conventional ``shallow" generative process is represented by a directed acyclic graph (DAG):
\begin{center}
\begin{tikzpicture}[var/.style={circle, draw=black, thick, minimum size=25pt}, boxvar/.style={rectangle, draw=black, thick, minimum size=28pt}, arrow/.style={-{Stealth[length=2.5mm]}, thick}]
\node[var] (Z) {$Z$};
\node[boxvar, right=2cm of Z] (X) {$X$};
\draw[arrow] (Z) -- (X);
\end{tikzpicture}
\end{center}
where we use circular nodes for latent variables and square nodes for observed variables. To create a deep structure, the DDM considers the following generative DAG:
\begin{center}
\begin{tikzpicture}[node distance=1.5cm, var/.style={circle, draw=black, thick, minimum size=28pt}, boxvar/.style={rectangle, draw=black, thick, minimum size=28pt}, dots/.style={minimum size=28pt}, arrow/.style={-{Stealth[length=2.5mm]}, thick}]
\node[var] (YT) {$Y_T$};
\node[var, right=of YT] (YT-1) {$Y_{T-1}$};
\node[dots, right=of YT-1] (dots) {$\cdots$};
\node[var, right=of dots] (Y1) {$Y_1$};
\node[boxvar, right=of Y1] (Y0) {$Y_0$};
\draw[arrow] (YT) -- (YT-1); \draw[arrow] (YT-1) -- (dots); \draw[arrow] (dots) -- (Y1); \draw[arrow] (Y1) -- (Y0);
\end{tikzpicture}
\end{center}
Here, we have a sequence of $T$ latent variables, where $Y_T = Z$ is pure noise and $Y_0 = X$ is the clean observation/image.

This Markovian structure implies that the joint PDF can be factorized as:
\begin{equation}
p(y_0,y_1,\dots, y_T) = p(y_T) p(y_{T-1}|y_T)\cdots p(y_0|y_1) = p(y_T)\prod_{t=0}^{T-1} p(y_t|y_{t+1}).
\label{eq::dag2}
\end{equation}
We assume the initial latent variable $Y_T$ follows a known distribution, such as a standard Gaussian, $p(y_T) = N(0, \mathbf{I}_d)$. The modeling effort then focuses on the conditional distributions for the reverse process, $p_{\theta_{t+1}}(y_t|y_{t+1})$. A common choice for these conditionals is a Gaussian parameterized by a neural network:
\begin{equation}
p_{\theta_{t+1}}(y_t|y_{t+1}) \sim N(\mu_{\theta_{t+1}}(y_{t+1}), \sigma^2_{\theta_{t+1}} (y_{t+1})\mathbf{I}_d),
\label{eq::G}
\end{equation}
where the full parameter set is $\theta = (\theta_1, \dots, \theta_{T})$. The joint PDF is therefore:
$$
p_\theta(y_0,y_1,\dots, y_T) = p(y_T)\prod_{t=0}^{T-1} p_{\theta_{t+1}}(y_t|y_{t+1}).
$$
The marginal log-likelihood for an observation $y_0$ requires integrating out all $T$ latent variables:
$$
\ell(\theta|y_0) = \log \int \dots \int p_\theta(y_0,y_1,\dots, y_T) dy_1\dots dy_T.
$$
Given data $X_1,\dots, X_n$, the MLE, $\hat \theta_{MLE} = {\sf argmax}_\theta \sum_{i=1}^n \ell(\theta|y_0 = X_i)$, is intractable.

As shown in Section \ref{sec::limit}, the EM algorithm fails even for a single layer of this model ($T=1$). With $T$ layers, the problem is significantly harder. To resolve this intractability, we again turn to variational approximation, specifically the AVI approach from Section \ref{sec::AVI}.

\subsection{Variational approximation}

To apply the AVI approach to the deep latent variable model, we first derive the corresponding ELBO:
\begin{equation}
\begin{aligned}
 \ell(\theta|y_0) & = \log \int p_\theta(y_0, y_1, \dots, y_T) dy_1\dots dy_T\\
 & =\log \int \frac{p_\theta(y_0, y_1, \dots, y_T)}{q_\phi(y_1,\dots, y_T|y_0)}q_\phi(y_1,\dots, y_T|y_0) dy_1\dots dy_T\\
 & \geq \int q_\phi(y_1,\dots, y_T|y_0) \log\left[\frac{p_\theta(y_0, y_1, \dots, y_T)}{q_\phi(y_1,\dots, y_T|y_0)}\right] dy_1\dots dy_T\\
 & = \mathbb{E}_{q_\phi}[\log p_\theta(y_0, Y_1, \dots, Y_T)] - \mathbb{E}_{q_\phi}[\log q_\phi(Y_1,\dots, Y_T|y_0)] \\
 & \equiv {\sf ELBO}_A(\theta, \phi|y_0),
\end{aligned}
\label{eq::DDM::ELBO}
\end{equation}
where $\mathbb{E}_{q_\phi}[\cdot]$ is the conditional mean of $Y_1,\cdots, Y_T$ given $Y_0=y_0$ under model $q_\phi$.
The challenge is to choose a tractable variational distribution $q_\phi(y_1,\dots, y_T|y_0)$. The Markovian structure of the generative model suggests a similar structure for the variational distribution. Specifically, we define the variational distribution or \emph{forward process} as a Markov chain proceeding from the observation $y_0$ to the final latent $y_T$:
\begin{equation}
q_\phi(y_1,\dots, y_T|y_0) = \prod_{t=0}^{T-1} q_{\phi_{t+1}}(y_{t+1}|y_{t}),
\label{eq::DDM::F1}
\end{equation}
where $\phi = (\phi_1,\dots, \phi_T)$. A convenient choice for these conditionals, which mirrors the Gaussian assumption of the reverse process, is:
\begin{equation}
q_{\phi_{t}}(y_{t}|y_{t-1}) \sim N( \sqrt{\phi_{t}} y_{t-1}, (1-\phi_{t}) \mathbf{I}_d),
\label{eq::DDM::F2}
\end{equation}
where each $\phi_t\in (0,1)$ is a variational parameter.
This corresponds to the Gaussian autoregressive-1 process:
\begin{equation}
Y_{t} = \sqrt{\phi_{t}} Y_{t-1} + \sqrt{1-\phi_{t} } E_{t},
\label{eq::DDM::F3}
\end{equation}
where $E_1,\dots, E_T$ are i.i.d. $N(0, \mathbf{I}_d)$. 
This process is easy to sample from. Moreover, for such Gaussian autoregressive model, we can sample $Y_t$ given $Y_0$ in one step:
\begin{equation}
\begin{aligned}
q_{\phi}(y_{t}|y_{0}) &\sim N\left(a_t y_0, b_t^2 \mathbf{I}_d\right),\\
a_t&= \sqrt{\prod_{s=1}^t\phi_s},\quad
b_t = \sqrt{1- \prod_{s=1}^t \phi_s}.
\end{aligned}
\label{eq::DDM::F4}
\end{equation}
This property is crucial for making the ELBO tractable. Substituting equation \eqref{eq::DDM::F1} into the ELBO and using the law of total expectation, we can decompose the ELBO into three main terms:
\begin{align*}
{\sf ELBO}_A(\theta, \phi|y_0) & = \underbrace{\mathbb{E}_{q_\phi}[\log p_{\theta_1}(y_0|Y_1)] + \sum_{t=1}^{T-1} \mathbb{E}_{q_\phi}[\log p_{\theta_{t+1}}(Y_t|Y_{t+1})]}_{=(A)} \\
 & \qquad+\underbrace{\mathbb{E}_{q_\phi}[\log p(Y_T)]}_{=(B)} - \underbrace{\sum_{t=0}^{T-1} \mathbb{E}_{q_\phi}[\log q_{\phi_{t+1}}(Y_{t+1}|Y_{t}) ]}_{=(C)}.
\end{align*}
Since the variational model is a Gaussian autoregressive process, terms (B) and (C) can be computed in closed form. Term (A) requires a Monte Carlo approximation, but this is made efficient by the one-shot sampling property of equation \eqref{eq::DDM::F4}. We now derive the analytical forms for (B) and (C).

\noindent\textbf{Term (B).} Assuming the prior $p(y_T) = N(0, \mathbf{I}_d)$, we have $\log p(y_T) = -\frac{d}{2}\log (2\pi) - \frac{1}{2}\|y_T\|^2$. Term (B) is then:
\begin{align*}
(B) = \mathbb{E}_{q_\phi}[\log p(Y_T)] &= -\frac{d}{2}\log (2\pi) - \frac{1}{2} \mathbb{E}_{Y_T\sim q_\phi(y_T|y_0)}[\|Y_T\|^2] \\
& \overset{\eqref{eq::DDM::F4}}{=} -\frac{d}{2}\log (2\pi)  -\frac{1}{2} [\|a_T y_0\|^2 + d b_T^2]\\
& =  -\frac{d}{2}\log (2\pi)  -\frac{1}{2} \left[\|y_0\|^2 \prod_{t=1}^T\phi_t + d \left(1- \prod_{t=1}^T \phi_t\right)\right].
\end{align*}

\noindent\textbf{Term (C).} Each term in the sum for (C) is the expected negative entropy of a conditional Gaussian:
\begin{align*}
\mathbb{E}_{q_\phi}[\log q_{\phi_{t+1}}(Y_{t+1}|Y_{t})] &=  \mathbb{E}_{q_\phi}\left[\mathbb{E}_{q_{\phi_{t+1}}(y_{t+1}|Y_t)}[\log q_{\phi_{t+1}}(Y_{t+1}|Y_{t})]\right]\\
& \overset{\eqref{eq::DDM::F2}}{=} \mathbb{E}_{q_\phi}\left[ -\frac{d}{2} \log(2\pi e (1-\phi_{t+1})) \right] \\
& = -\frac{d}{2} \log(2\pi e (1-\phi_{t+1})),
\end{align*}
where we use the fact that the negative entropy of $N(\cdot, (1-\phi_{t+1})\mathbf{I}_d)$ is $ -\frac{d}{2} \log(2\pi e (1-\phi_{t+1}))$
and $\mathbb{E}_{q_{\phi_{t+1}}(y_{t+1}|Y_t)}[\cdot]$ refers to conditional mean of $Y_{t+1}$ given $Y_{t} $  under model $q_{\phi_{t+1}}$.
Summing over all terms:
$$
(C) = -\frac{dT}{2} \log\left(2\pi e\right) - \frac{d}{2} \sum_{t=1}^T \log\left(1-\phi_{t}\right).
$$
Dropping terms irrelevant to $\theta$ and $\phi$, we obtain a refined ELBO for optimization:
\begin{equation}
\begin{aligned}
{\sf ELBO}^*_A(\theta, \phi|y_0) 
 &= \mathbb{E}_{q_\phi}[\log p_{\theta_1}(y_0|Y_1)] + \sum_{t=2}^{T} \mathbb{E}_{q_\phi}[\log p_{\theta_{t}}(Y_{t-1}|Y_{t})]\\
 & \quad -\frac{1}{2} \|y_0\|^2 \prod_{t=1}^T\phi_t - \frac{d}{2}\left(1- \prod_{t=1}^T \phi_t\right) -\frac{d}{2} \sum_{t=1}^T \log (1-\phi_t).
\end{aligned}
\label{eq::ELBO::DDM}
\end{equation}
Given data $X_1,\dots, X_n$, the estimators for the DDM are found by maximizing the total ELBO:
$$
(\hat \theta_{DDM}, \hat \phi) = {\sf argmax}_{\theta, \phi} \sum_{i=1}^n {\sf ELBO}^*_A(\theta, \phi|y_0 = X_i).
$$

\subsection{Gradient of the DDM's ELBO} \label{sec::DDM::GD}

Since the DDM is a special case of the AVI/VAE framework, the gradient computation follows the same principles outlined in Section \ref{sec::GD::AVI} and Appendix \ref{sec::G::AVI}. 
Note that in standard DDM implementations \citep{ho2020ddpm}, the variational parameters $\phi_1, \dots, \phi_T$ are not learned. Instead, they are pre-defined as a fixed hyperparameter. This simplifies the optimization to be solely over the generative model parameters; see Section \ref{sec::practical} for more discussion.
However, variational parameters $\phi_1, \dots, \phi_T$ are learnable if needed.
The DDM's forward process is, by construction, a Gaussian autoregressive model, so the reparameterization trick is directly applicable for computing gradients with respect to the variational parameters $\phi$.

\begin{figure}
\centering
\includegraphics[height=1.25in]{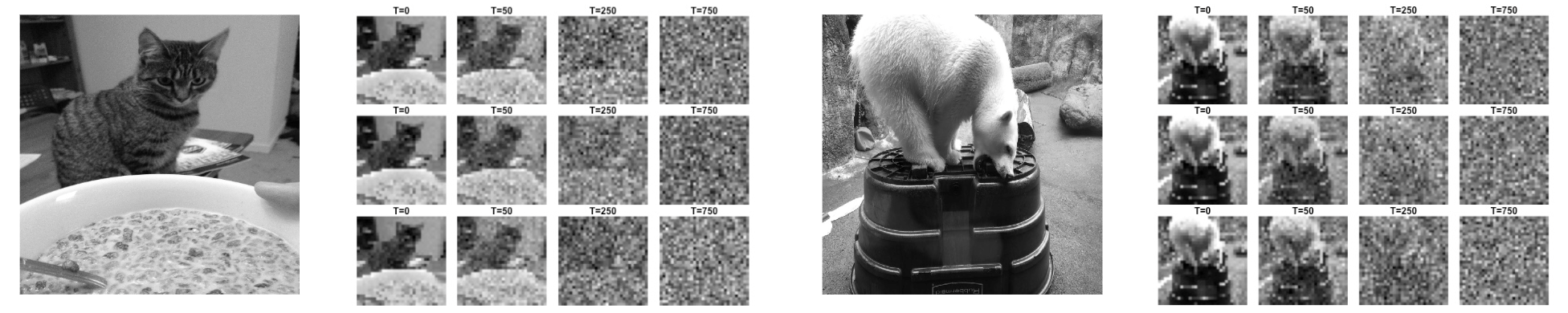}
\caption{An illustration of the DDM training loop. For each observation (e.g., the cat and polar bear images), the forward process is used to generate a trajectory of noisy images. These trajectories are then used to compute the gradients and update the generative model's parameters in the reverse (denoising) process. This is repeated until convergence.}
\label{fig::DDM::VI2}
\end{figure}

The gradient of the refined ELBO with respect to the generative model parameters $\theta$ is separable for each parameter $\theta_t$:
\begin{equation}
\begin{aligned}
\nabla_{\theta_t}{\sf ELBO}^*_A(\theta, \phi|y_0) = 
\begin{cases}
\mathbb{E}_{q_\phi}[\nabla_{\theta_1} \log p_{\theta_1}(y_0|Y_1)], & \text{for } t=1 \\
\mathbb{E}_{q_\phi}[\nabla_{\theta_t} \log p_{\theta_{t}}(Y_{t-1}|Y_{t})], & \text{for } t=2,\dots, T.
\end{cases}
\end{aligned} 
\label{eq::DDM::G}
\end{equation}
The Monte Carlo approximation for this gradient involves generating full latent trajectories. We first sample a sequence
$$
\tilde{\mathbf{Y}} = (\tilde Y_0 = y_0, \tilde Y_1, \tilde Y_2,\dots, \tilde Y_T)
$$
by applying the forward process (equation \eqref{eq::DDM::F3}) iteratively. By repeating this $M$ times, we obtain $M$ independent trajectories, $\tilde{\mathbf{Y}}^{(1)},\dots, \tilde{\mathbf{Y}}^{(M)}$. With these samples, we can approximate the expectation in equation \eqref{eq::DDM::G} as:
\begin{equation}
\begin{aligned}
\tilde{\nabla_{\theta_t}{\sf ELBO}^*_A}(\theta, \phi|y_0) &\approx \frac{1}{M}\sum_{m=1}^M\nabla_{\theta_t} \log p_{\theta_{t}}(\tilde{Y}^{(m)}_{t-1}|\tilde{Y}^{(m)}_{t}) \\
&= \frac{1}{M}\sum_{m=1}^M s(\theta_t|\tilde{Y}^{(m)}_{t-1},\tilde{Y}^{(m)}_{t}),
\end{aligned}
\label{eq::DDM::G2}
\end{equation}
where $s(\theta_t|y_{t-1},y_{t}) = \nabla_{\theta_t} \log p_{\theta_{t}}(y_{t-1}|y_{t})$ is the score function of the conditional generative model at step $t$. Figure \ref{fig::DDM::VI2} provides a graphical illustration of this training process.

{\bf Data-generating process as a `denoising' process.}
The form of the gradient in equation \eqref{eq::DDM::G2} provides a crucial insight. The learning signal for the parameter $\theta_t$ comes from the score function of $p_{\theta_{t}}(y_{t-1}|y_{t})$. This task is effectively asking the model to predict a cleaner state, $\tilde{Y}_{t-1}$, given a noisier one, $\tilde{Y}_{t}$. Thus, the generative (reverse) model $p_\theta$ learns to progressively \emph{denoise} a sequence of latent variables, starting from pure noise $Y_T$ and ending at a clean image $Y_0$. 
\subsection{Forward and reverse processes}

The variational framework described above casts the DDM as a specific type of VAE. The \emph{decoder} is our data-generating model, $p_\theta$, which describes how to generate an observation $Y_0$ from a pure noise variable $Y_T = Z$. The \emph{encoder} is our variational distribution, $q_\phi$, which is a Gaussian autoregressive model.
In the DDM literature, these two components are known as the forward and reverse processes.

The encoder, $q_\phi$, which maps the observation $Y_0$ to the final latent noise variable $Y_T$, is called the \textbf{forward process}. It is a Gaussian autoregressive model that sequentially adds Gaussian noise to the observation (as in equation \eqref{eq::DDM::F3}), which is analogous to a diffusion process.

The decoder, $p_\theta$, operates in the opposite direction. It starts with pure noise $Y_T$ and sequentially removes the noise to recover the original observation $Y_0$. This is called the \textbf{reverse process} and is functionally a denoising process. The combination of these two components gives the Denoising Diffusion Model its name.

Many tutorials on DDMs begin by introducing the forward process before deriving the reverse process \citep{ho2020ddpm,luo2022understanding} since this aligns with the implementation--the computer will perform forward process first and then use the reverse process to fit the parameter $\theta$. This contrasts with the statistical modeling tradition, which typically begins from the data-generating model (the reverse process) and then constructs the variational approximation (the forward process) as a tool for tractable inference. 

To summarize the parallel terminologies:
\begin{itemize}
    \item \textbf{Decoder = Reverse Process = Data-Generating Model:} A deep latent variable model with a Markov chain structure that learns to progressively denoise a variable from pure noise into an observation.
    \item \textbf{Encoder = Forward Process = Variational Distribution:} A Gaussian autoregressive model with a similar Markov structure that progressively adds noise to an observation.
\end{itemize}

\subsection{Practical implementation and the simplified objective}	\label{sec::practical}

The full ELBO provides the theoretical foundation for DDMs, but in practice, practitioners have adopted several key specifications to yield a more stable and efficient objective function, enabling large-scale training.

{\bf Fixing the variational parameter and covariance matrix model. }
In practice, the DDM training process is made more efficient through several key specifications. First, the parameters of the variational distribution (the forward process) are not learned from data. Instead, they are fixed as pre-defined hyperparameters, collectively known as the \emph{variance schedule} \citep{ho2020ddpm}.
Moreover, the covariance matrix in the reverse (data-generating) process is also assumed to be fixed and diagonal, typically as $\Sigma_{\theta_t}(y_t) = \sigma_t^2\mathbf{I}_d$. The variances $\sigma_t^2$ are known constants, often tied to the forward process variance schedule.
This specification has two main benefits. First, it removes the need to learn any variance parameters. Second, it simplifies the part of the ELBO related to $\theta$ into a weighted least-squares objective. As shown in Equation \eqref{eq::DDM::G}, the gradient of the ELBO with respect to the mean function $\mu_{\theta_t}$ becomes:
\begin{equation}
\begin{aligned}
\mathbb{E}_{q_\phi}[\nabla_\theta \log p_{\theta_{t}}(Y_{t-1}|Y_{t})] 
&= \mathbb{E}_{q_\phi}\left[\nabla_\theta \left(\frac{-1}{2\sigma_t^2} \|Y_{t-1} - \mu_{\theta_t}(Y_t)\|^2\right) \right] \\
& = -\frac{1}{2\sigma_t^2} \mathbb{E}_{q_\phi}\left[\nabla_\theta \|Y_{t-1} - \mu_{\theta_t}(Y_t)\|^2 \right].
\end{aligned}
\label{eq::DDM::S1}
\end{equation}
The optimization, therefore, reduces to training the model $\mu_{\theta_t}$ to predict the mean of the denoised variable $Y_{t-1}$. This gradient can be efficiently estimated using samples generated from the fixed forward process.

{\bf Shared parameters in the generative model $p_\theta$.}
Moreover, to further reduce model complexity, 
people often harmonize the models $p_{\theta_1}(y_0|y_1),\cdots, p_{\theta_T}(y_{T-1}|y_T)$ so that 
instead of using different parameters for each conditional model, they use the same shared parameter but include the step $t$ as a covariate.
Specifically, the new conditional model is
\begin{equation}
p_{\theta}(y_{t-1}|y_t) \sim N\left(\mu_\theta(y_t, t), \sigma_t^2\mathbf{I}_d\right).
\label{eq::DDM::S2}
\end{equation}
With this, the gradient in equation \eqref{eq::DDM::S1} is updated to 
\begin{equation}
\begin{aligned}
\mathbb{E}_{q_\phi}[\nabla_\theta \log p_{\theta_{t}}(Y_{t-1}|Y_{t})] 
& = \frac{1}{2\sigma_t^2} \mathbb{E}_{q_\phi}\left[\nabla_\theta \|Y_{t-1} - \mu_{\theta}(Y_t, t)\|^2 \right].
\end{aligned}
\label{eq::DDM::S3}
\end{equation}
This greatly reduces the model complexity.

\subsubsection{Noise prediction formulation}	\label{sec::noise}
The key insight from \cite{ho2020ddpm} is that this objective can be re-written as a noise prediction task. 
The key criterion of equation \eqref{eq::DDM::S3} is the expectation (moving the gradient operator $\nabla_\theta$ out for simplicity)
\begin{equation}
\begin{aligned}
\mathbb{E}_{q_\phi}\left[ \|Y_{t-1} - \mu_{\theta}(Y_t, t)\|^2 \right]
& = \int \|y_{t-1} - \mu_{\theta}(y_t, t)\|^2 q_\phi(y_t,y_{t-1}|y_0)dy_tdy_{t-1}\\
& = \int \int \|y_{t-1} - \mu_{\theta}(y_t, t)\|^2 q_\phi(y_{t-1}|y_{t}, y_0)dy_{t-1} q_{\phi_t}(y_t|y_{0})dy_t\\
& = \int \E_{q_\phi(y_{t-1}|y_t,y_0)} \left[\|Y_{t-1} - \mu_{\theta}(y_t, t)\|^2 \right] q_{\phi_t}(y_t|y_{0})dy_t,
\end{aligned}
\label{eq::DDM::S4}
\end{equation}
where $\E_{q_\phi(y_{t-1}|y_t,y_0)}[\cdot]$ refers to the conditional mean of $Y_{t-1}$ given $Y_t=y_t, Y_0 = y_0$ under $q_\phi$.

By equations \eqref{eq::DDM::F3} and \eqref{eq::DDM::F4},
the conditional distribution $q_\phi(y_{t-1}|y_{t}, y_0)$ is $N(\tilde{\mu}_t(y_{t},y_0), \tilde{\sigma}_t^2\mathbf{I}_d)$ such that
\begin{equation}
\begin{aligned}
\tilde{\mu}_t(y_{t},y_0)  &= \frac{\sqrt{\phi_t }(1-\prod_{s=1}^{t-1}\phi_s )}{1-\prod_{s=1}^{t}\phi_s }y_t + \frac{\sqrt{\prod_{s=1}^{t-1} \phi_s} (1-\phi_t)}{1-\prod_{s=1}^{t}\phi_s}y_0\\
& = \frac{\sqrt{\phi_t} b^2_{t-1}}{b_t^2} y_t + \frac{a_{t-1}(1-\phi_t)}{b_t^2} y_0, \\
\tilde{\sigma}_t^2 &= \frac{(1-\phi_t)(1-\prod_{s=1}^{t-1}\phi_s)}{1-\prod_{s=1}^t \phi_s}\\
& = \frac{(1-\phi_t)b_{t-1}^2}{b_t^2},
\end{aligned}
\label{eq::DDM::S5}
\end{equation}
where $a_t= \sqrt{\prod_{s=1}^t\phi_s} $ and $b_t = \sqrt{1- \prod_{s=1}^t \phi_s}$.
Thus,
the inner expectation in equation \eqref{eq::DDM::S4}, 
$$
\E_{q_\phi(y_{t-1}|y_t,y_0)} \left[\|Y_{t-1} - \mu_{\theta}(y_t, t)\|^2 \right],
$$ 
is in the form of $\E[\|W - \zeta_\theta\|^2] = C + \|\mu_W - \zeta_\theta\|^2 $, where $W \sim N(\mu_W, \Sigma_W)$ and $C$ is a constant independent of $\theta$.
So we have 
$$
\E_{q_\phi(y_{t-1}|y_t,y_0)} \left[\|Y_{t-1} - \mu_{\theta}(y_t, t)\|^2 \right] = C_1 + \|\tilde{\mu}_t(y_{t},y_0) - \mu_{\theta}(y_t, t)\|^2
$$
for some constant $C_1$.
Therefore, based on equation \eqref{eq::DDM::S5},  equation \eqref{eq::DDM::S4} can be rewritten as 
\begin{equation}
\begin{aligned}
\mathbb{E}_{q_\phi}\left[ \|Y_{t-1} - \mu_{\theta}(Y_t, t)\|^2 \right]
& = C_1+\int \|\tilde{\mu}_t(y_{t},y_0) - \mu_{\theta}(y_t, t)\|^2q_{\phi_t}(y_t|y_{0})dy_t\\
& = C_1 + \int \left\|\frac{\sqrt{\phi_t} b^2_{t-1}}{b_t^2} y_t + \frac{{a_{t-1}} (1-\phi_t)}{b_t^2} y_0 -  \mu_{\theta}(y_t, t)\right\|^2q_{\phi_t}(y_t|y_{0})dy_t.
\end{aligned}
\label{eq::DDM::S6}
\end{equation}
Because $q_{\phi}(y_{t}|y_{0}) \sim N\left(a_t y_0, b_t^2 \mathbf{I}_d\right)$,
we can express $Y_t$ as a function of $y_0$ and the isotropic Gaussian noise $E \sim N(0, \mathbf{I}_d)$ 
as 
$Y_t = a_t y_0 + b_t E$.
This means that $y_0 = \frac{1}{a_t} (Y_t - b_t E)$. 
With this, we can rewrite the integrand in equation \eqref{eq::DDM::S6} as 
\begin{equation}
\begin{aligned}
\Bigg\|\frac{\sqrt{\phi_t} b^2_{t-1}}{b_t^2} y_t &+ \frac{{a_{t-1}} (1-\phi_t)}{b_t^2} y_0 -  \mu_{\theta}(y_t, t)\Bigg\|^2\\
& = \left\|\frac{\sqrt{\phi_t} b^2_{t-1}}{b_t^2} y_t + \frac{{a_{t-1}} (1-\phi_t)}{b_t^2} \frac{1}{a_t}(y_t - b_t e)-  \mu_{\theta}(y_t, t)\right\|^2\\
& = \left\|\frac{1}{\sqrt{\phi_t}} y_t -\frac{1-\phi_t}{\sqrt{\phi_t} b_t} e - \mu_{\theta}(y_t, t)\right\|^2\\
& = \left\|\frac{1-\phi_t}{\sqrt{\phi_t} b_t} \Psi_\theta(y_t, t) -\frac{1-\phi_t}{\sqrt{\phi_t} b_t} e \right\|^2\\
& = \frac{(1-\phi_t)^2}{{\phi_t} b^2_t}\left\| \Psi_\theta(y_t, t) - e \right\|^2\\
& = \frac{(1-\phi_t)^2}{{\phi_t} b^2_t}\left\| \Psi_\theta(a_t y_0 + b_t e, t) - e \right\|^2,
\end{aligned}
\label{eq::DDM::S7}
\end{equation}
where $e$ is the variable corresponding to random noise $E$ and
we rewrite the model $\mu_\theta$ as 
\begin{equation}
\begin{aligned}
\mu_\theta(y_t,t) &= \frac{1}{\sqrt{\phi_t}} y_t - \frac{1-\phi_t}{\sqrt{\phi_t} b_t} \Psi_\theta(y_t,t).
\end{aligned}
\label{eq::DDM::S8}
\end{equation}
In this construct,
$\Psi_\theta(y_t,t) = \frac{b_t}{1-\phi_t}(\sqrt{\phi_t}  \mu_\theta(y_t,t) - y_t)$ is just a rescaled version of $\mu_\theta$,
so learning the parameter $\theta$ using $\Psi_\theta(y_t,t)$ is the same as $\mu_\theta(y_t,t)$ when
the variational parameters $\phi_t, b_t$ are fixed.

By equations \eqref{eq::DDM::S7} and \eqref{eq::DDM::S8}, we can rewrite the expectation in equation \eqref{eq::DDM::S6} as 
\begin{equation}
\begin{aligned}
\mathbb{E}_{q_\phi}\left[ \|Y_{t-1} - \mu_{\theta}(Y_t, t)\|^2 \right]
& = C_1+\int \|\tilde{\mu}_t(y_{t},y_0) - \mu_{\theta}(y_t, t)\|^2q_{\phi_t}(y_t|y_{0})dy_t\\
& = C_1 + \frac{(1-\phi_t)^2}{{\phi_t} b^2_t} \int \left\| \Psi_\theta(a_t y_0 + b_t e, t) - e \right\|^2 p_E(e) de\\
& = C_1 +\frac{(1-\phi_t)^2}{{\phi_t} b^2_t} \E_{E\sim p_E}\left[ \left\| \Psi_\theta( a_t y_0 + b_t E, t) - E \right\|^2\right],
\end{aligned}
\label{eq::DDM::S9}
\end{equation}
where $p_E(e)$ is the PDF of $ N(0, \mathbf{I}_d)$ and $E\sim p_E$. 
Equation \eqref{eq::DDM::S9} shows an interesting interpretation about the model $\Psi_\theta( a_t y_0 + b_t e, t) $. 
This model predicts the added noise $e$ to the original observation/image $y_0$.
So our reverse process is using the learned parameter to denoise $Y_T$ back to the original observation. 

In summary, under the following model specifications:
\begin{itemize}
\item {\bf Fixed variational parameters.} $\phi_1,\cdots, \phi_T$ are fixed,
\item {\bf Fixed covariance matrix.} The covariance matrix $\Sigma_\theta(y_t) = \sigma_t^2$ is fixed,
\item {\bf Shared parameters in the generative model $p_\theta$. } The mean function in the generative model $\mu_{\theta_t}(y_t) = \mu_\theta(y_t, t)$,
\end{itemize}
the ELBO in equation \eqref{eq::ELBO::DDM} can be expressed as 
\begin{equation}
\begin{aligned}
{\sf ELBO}^*_A(\theta, \phi|y_0) & = {\sf ELBO}^*_A(\theta|y_0)\\
& = - \sum_{t=1}^T\frac{(1-\phi_t)^2}{ \sigma_t^2{\phi_t} b^2_t} \E\left[ \left\| \Psi_\theta(a_t y_0 + b_t E, t) - E \right\|^2 \right] + C_2
\end{aligned}
\label{eq::DDM::ELBO1}
\end{equation}
for some constant $C_2$.
Thus, maximizing ${\sf ELBO}^*_A(\theta, \phi|y_0)$ is equivalent to minimizing 
\emph{the square errors}:
$$
 \sum_{t=1}^T\frac{(1-\phi_t)^2}{ \sigma_t^2{\phi_t} b^2_t} \E\left[ \left\| \Psi_\theta(a_t y_0 + b_t E, t) - E \right\|^2 \right].
 $$

Equation \eqref{eq::DDM::ELBO1} can be numerically computed easily   since $E\sim N(0,\mathbf{I}_d)$,
so the gradient 
$$
\nabla_\theta  {\sf ELBO}^*_A(\theta|y_0) = - \sum_{t=1}^T\frac{(1-\phi_t)^2}{ \sigma_t^2{\phi_t} b^2_t} \E\left[ \nabla_\theta\left\| \Psi_\theta(a_t y_0 + b_t E, t) - E \right\|^2 \right]
$$
can be approximated efficiently via a Monte Carlo method.

\cite{ho2020ddpm} found that empirically, ignoring the multiplicative factor $\frac{(1-\phi_t)^2}{ \sigma_t^2{\phi_t} b^2_t} $
did not change the result much, so they proposed to learn $\theta$ via minimizing 
$$
 \sum_{t=1}^T \E\left[ \nabla_\theta\left\| \Psi_\theta(a_t y_0 + b_t E, t) - E \right\|^2 \right]
$$
and introduced a stochastic method that replace $\sum_{t=1}^T$ by a random number. 
Specifically, we generate
$$
\tilde{U}^{(1)},\cdots, \tilde{U}^{(M)} \sim {\sf Uni}\{1,2,\cdots, T\}, \quad \tilde{E}^{(1)},\cdots, \tilde{E}^{(M)}\sim N(0,\mathbf{I}_d)
$$
and approximate the gradient of ELBO for observation $y_0$ as
\begin{equation}
\tilde{\nabla_\theta{\sf ELBO}^*_A}(\theta|y_0) =  \frac{1}{M}\sum_{i=1}^m\nabla_\theta\left\| \Psi_\theta(a_{\tilde {U}^{(m)}} y_0 + b_{\tilde{U}^{(m)}} \tilde{E}^{(m)}, \tilde{U}^{(m)}) - \tilde{E}^{(m)} \right\|^2.
\label{eq::DDM::ELBO2}
\end{equation}
The gradient in equation \eqref{eq::DDM::ELBO2} is a lot easier to compute
than the gradient in equation \eqref{eq::DDM::G2}
because we no longer need to run the entire forward process. Instead, we just need to generate 
a lot of random integers $\tilde{U}^{(m)}\in\{1,2,\cdots, T\}$ and isotropic Gaussians $\tilde{E}^{(m)}\sim N(0,\mathbf{I}_d)$
to learn the parameter $\theta$.

\section{Conclusion}

Variational inference (VI), variational autoencoders (VAEs), and diffusion models (DDMs) share a common foundation in 
\emph{latent variable modeling and likelihood approximation}. 
Starting from the classical EM algorithm, we have seen that VI arises as a natural relaxation of the intractable E-step 
by replacing the conditional distribution $p(z|x=X_i;\theta^{(t)})$ with a tractable variational family $q_{\omega_i}(z)$. 
Amortized VI further simplifies computation by learning a conditional mapping $q_\phi(z|x)$, 
enabling large-scale estimation and forming the backbone of VAEs. 
Finally, the DDM extends this framework into a 
\emph{deep latent variable model} with a Markov chain structure, 
providing one of the most powerful modern generative modeling tools.

\subsection{Variational inference: Frequentist or Bayesian?}

While VI is often introduced as a Bayesian approach 
\citep{blei2017variational,doersch2016vae,kingma2014vae}, 
it is not inherently Bayesian. 
In our analysis, VI was developed entirely from a \emph{frequentist} perspective: 
we did not place any prior on the parameter of interest~$\theta$. 
Instead, VI served purely as a computational device for approximating 
the maximum likelihood estimator when the likelihood is intractable.

That said, VI can also be viewed in a Bayesian context if the primary target of inference 
is the latent variable~$Z$ rather than the model parameter~$\theta$\footnote{Another common Bayesian setting 
is that we place a prior distribution on $(\theta,z)$ and use variational inference to approximate $p(\theta,z|x).$}.
In that case, the distribution~$p(z)$ plays the role of a prior, 
and the intractable conditional~$p(z | x; \theta)$ represents the posterior distribution. 
The variational distributions~$q_\omega(z)$ or~$q_\phi(z | x)$ 
then provide tractable approximations to this posterior.

Ultimately, VI is best understood as a general computational framework 
for approximating intractable conditional distributions~$p(z | x; \theta)$. 
It applies equally well to frequentist settings, such as latent space models, 
and to Bayesian problems, such as posterior inference on latent variables. 
From either perspective, VI unifies computational tractability and probabilistic approximation 
through the same underlying optimization principle.

\subsection{Latent variable modeling: generative utility versus scientific interpretability}	\label{sec::LVM}

The role of latent variables in deep generative models (VAEs, DDMs) diverges sharply from their role in traditional statistics--it is a distinction between generative utility and scientific interpretability.

In VAEs and DDMs, latent variables serve primarily as a tool to construct flexible, high-capacity models capable of approximating complex data distributions, such as those of natural images. The principal objective is generative performance--producing realistic data--with computational tractability as a key constraint. Consequently, the interpretability of individual latent dimensions is often secondary, and model architecture is freely modified to improve results. The model specification of DDMs that enables the noise prediction formulation (Section \ref{sec::practical}) highlights this principle.

Conversely, in classical latent variable methods like factor analysis, the primary goal is scientific interpretation \citep{anderson2003introduction,harman1976modern}. Latent variables are hypothesized to represent meaningful, underlying constructs rooted in domain knowledge. Their meaning is paramount, and any change to the model's latent structure requires strong theoretical or statistical justification. Thus, despite procedural similarities, the two paradigms are guided by different philosophies: one driven by predictive power, the other by explanatory insight.

\bibliography{EMVar}

\bibliographystyle{abbrvnat}

\appendix
\section{Gradient of the Amortized ELBO} \label{sec::G::AVI}

\subsection{Gradient with respect to model parameters \texorpdfstring{$\theta$}{theta}}
The gradient with respect to the model parameters $\theta$ is generally straightforward to compute. Recall the amortized ELBO:
$$
{\sf ELBO}_A(\theta,\phi|x)  = \int q_\phi(z|x) \ell(\theta|x,z) dz - \int q_{\phi}(z|x) \log q_\phi(z|x)dz.
$$
Only the first term depends on $\theta$, so the gradient is:
\begin{align*}
\nabla_\theta{\sf ELBO}_A(\theta,\phi|x) & = \nabla_\theta \int q_\phi(z|x) \ell(\theta|x,z) dz \\
& = \int q_\phi(z|x) s(\theta|x,z) dz = \mathbb{E}_{Z\sim q_\phi(\cdot|x)}[s(\theta|x,Z)],
\end{align*}
where $s(\theta|x,z) = \nabla_\theta \ell(\theta|x,z)$ is the complete-data score function. Because the variational distribution $q_\phi(z|x)$ is designed to be easy to sample from, we can efficiently approximate this gradient via a Monte Carlo method. We generate $\tilde z^{(1)},\dots, \tilde z^{(M)}$ from $q_\phi(z|x)$ and compute the estimate:
\begin{equation}
\tilde{\nabla_\theta{\sf ELBO}_A}(\theta,\phi|x) = \frac{1}{M} \sum_{m=1}^M s(\theta|x,\tilde z^{(m)}).
\label{eq::ELBO::theta}
\end{equation}

\subsection{Gradient with respect to variational parameters \texorpdfstring{$\phi$}{phi} and the reparameterization trick}
As with non-amortized VI, the reparameterization trick is applicable when the variational distribution is Gaussian, providing an efficient path to numerical optimization. We assume here that $q_\phi(z|x)$ is a multivariate Gaussian, $N(\eta_\phi(x), \Omega_\phi(x))$, where the mean function $\eta_\phi(x)$ and covariance function $\Omega_\phi(x)$ are parameterized by $\phi$. Let $L_\phi(x)$ be the Cholesky decomposition of the covariance matrix, such that $\Omega_\phi(x) = L_\phi(x)L_\phi(x)^T$. A sample $Z \sim q_\phi(\cdot|x)$ can be reparameterized as:
\begin{equation}
Z = \eta_\phi(x) + L_\phi(x) \epsilon,\qquad \text{where } \epsilon \sim N(0, \mathbf{I}_k).
\label{eq::Rep}
\end{equation}
The gradient of the ELBO with respect to $\phi$ consists of two terms:
\begin{equation}
\nabla_\phi {\sf ELBO}_A(\theta,\phi|x) = \nabla_\phi\int q_\phi(z|x) \ell(\theta|x,z) dz + \nabla_\phi\left(-\int q_{\phi}(z|x) \log q_\phi(z|x)dz\right).
\label{eq::ELBO::phi}
\end{equation}
The second term is the gradient of the entropy. For a Gaussian, the negative entropy has an analytical form: $-\frac{k}{2}\log (2\pi e) - \frac{1}{2}\log {\sf det}(\Omega_\phi(x))$. Its gradient is therefore:
\begin{equation}
-\nabla_\phi\left(\int q_{\phi}(z|x) \log q_\phi(z|x)dz\right) = -\frac{1}{2}\nabla_\phi  \log {\sf det}(\Omega_\phi(x)),
\label{eq::phi}
\end{equation}
which typically has a closed-form expression once the structure of $\Omega_\phi(x)$ is specified.

The main challenge is the first term, where the derivative is with respect to the parameters of the sampling distribution. The reparameterization trick (equation \eqref{eq::Rep}) resolves this by rewriting the integral as an expectation over the fixed distribution of $\epsilon$:
\begin{align*}
\nabla_\phi  \int q_\phi(z|x) \ell(\theta|x,z) dz & = \nabla_\phi  \mathbb{E}_{\epsilon \sim N(0,\mathbf{I}_k)}[\ell(\theta|x,\eta_\phi(x) + L_\phi(x)\epsilon)] \\
&=   \mathbb{E}_{\epsilon \sim N(0,\mathbf{I}_k)}[\nabla_\phi \ell(\theta|x,\eta_\phi(x) + L_\phi(x)\epsilon)]\\
& = \mathbb{E}_{\epsilon \sim N(0,\mathbf{I}_k)}\left[ (\nabla_\phi z) \cdot \nabla_z \ell(\theta|x,z)|_{z=\eta_\phi(x) + L_\phi(x)\epsilon} \right],
\end{align*}
where $\nabla_\phi z = \nabla_\phi \eta_\phi(x) + (\nabla_\phi L_\phi(x))\epsilon$. This expectation can be estimated via Monte Carlo. We generate $\tilde{\epsilon}^{(1)},\dots, \tilde{\epsilon}^{(M)}\sim N(0, \mathbf{I}_k)$ and compute:
$$
\tilde{\nabla_\phi\mathbb{E}_{Z|X=x\sim q_\phi}}[\ell(\theta|x,z)] =\frac{1}{M}\sum_{m=1}^M \left[\nabla_\phi \eta_\phi(x) + (\nabla_\phi L_\phi(x))\tilde{\epsilon}^{(m)}\right]\nabla_z\ell(\theta|x,\eta_\phi(x) + L_\phi(x)\tilde{\epsilon}^{(m)}).
$$
Combining the Monte Carlo estimate for the first term with the analytical gradient of the entropy term, the full gradient of the ELBO with respect to $\phi$ is estimated as:
\begin{equation}
\tilde{\nabla_\phi {\sf ELBO}_A}(\theta,\phi|x) = \widehat{\nabla_\phi\mathbb{E}_{Z|X=x\sim q_\phi}}[\ell(\theta|x,Z)] - \frac{1}{2}\nabla_\phi  \log {\sf det}(\Omega_\phi(x)).
\label{eq::ELBO::phi2}
\end{equation}
The gradient estimates from equations \eqref{eq::ELBO::theta} and \eqref{eq::ELBO::phi2} are then used in the gradient ascent procedure (equation \eqref{eq::GD}) to numerically compute the estimators $\hat\theta_{AVI}$ and $\hat\phi$.

\end{document}